\documentclass[11pt,a4paper]{article}
\usepackage[hyperref]{acl2020}
\usepackage{times}
\usepackage{latexsym}

\usepackage{natbib} % has a nice set of citation styles and commands

\usepackage{graphicx}
\usepackage[american]{babel}
\usepackage{mathtools} % amsmath with fixes and additions
\usepackage{booktabs} % commands to create good-looking tables
\usepackage{tikz} % nice language for creating drawings and diagrams
\usepackage{dsfont}
\usepackage{multirow} 
\usepackage{makecell}
\usepackage{tabularx}
\usepackage{booktabs}
\usepackage{array,mathbbol}
\usepackage{bm}
\usepackage{amsfonts,amssymb}
\usepackage{amsmath,graphicx}
\usepackage{float}
\usepackage{csquotes}
\usepackage{wrapfig}
\usepackage{balance}
\graphicspath{ {./img/} }

\makeatletter
\renewenvironment*{displayquote}
  {\begingroup\setlength{\leftmargini}{0cm}\csq@getcargs{\csq@bdquote{}{}}}
  {\csq@edquote\endgroup}
\makeatother

\usepackage{etoolbox}
\newcommand{\zerodisplayskips}{
  \setlength{\abovedisplayskip}{0pt}
  \setlength{\belowdisplayskip}{8pt}
  \setlength{\abovedisplayshortskip}{0pt}
  \setlength{\belowdisplayshortskip}{0pt}
}
\appto{\normalsize}{\zerodisplayskips}
\appto{\small}{\zerodisplayskips}
\appto{\footnotesize}{\zerodisplayskips}

\usepackage{microtype}

\aclfinalcopy % Uncomment this line for the final submission
\def\aclpaperid{34} %  Enter the acl Paper ID here

\title{End-to-End Annotator Bias Approximation \\ on Crowdsourced Single-Label Sentiment Analysis}

\author{
Gerhard Hagerer \and
David Szabo \and
Andreas Koch \and 
Maria Luisa Ripoll Dominguez \and \\
{\bf Christian Widmer} \and
{\bf Maximilian Wich} \and
{\bf Hannah Danner} \and
{\bf Georg Groh} \\
Technical University of Munich, Germany \\
\texttt{\{ghagerer, grohg\}@mytum.de}
}

\date{}

\begin{document}
\maketitle
\begin{abstract}
Sentiment analysis is often a crowdsourcing task prone to subjective labels given by many annotators. It is not yet fully understood how the annotation bias of each annotator can be modeled correctly with state-of-the-art methods. However, resolving annotator bias precisely and reliably is the key to understand annotators' labeling behavior and to successfully resolve corresponding individual misconceptions and wrongdoings regarding the annotation task. Our contribution is an explanation and improvement for precise neural end-to-end bias modeling and ground truth estimation, which reduces an undesired mismatch in that regard of the existing state-of-the-art. Classification experiments show that it has potential to improve accuracy in cases where each sample is annotated only by one single annotator. We provide the whole source code publicly\footnote{\url{https://github.com/theonlyandreas/end-to-end-crowdsourcing}} and release an own domain-specific sentiment dataset containing 10,000 sentences discussing organic food products\footnote{\url{https://github.com/ghagerer/organic-dataset}}. These are crawled from social media and are singly labeled by 10 non-expert annotators.
\end{abstract}

\section{Introduction}

%The world of today is marked by movements for equality intending to reduce potentially offending biases that have previously been part of the fabric of society. Given these debates on social equality, science has followed this trend, as the topics of ethical AI and machine learning bias gain relevance. More and more datasets offer annotator information to help detect undesired prejudices and discrimination caused by race, gender, age, and other traits \cite{thelwall2018gender,kiritchenko2018examining}. 

Modeling annotator bias in conditions where each data point is annotated by multiple annotators, below referred to as multi-labeled crowdsourcing, has been investigated thoroughly. However, bias modeling when every data point is annotated by only one person, hereafter called singly labeled crowdsourcing, poses a rather specific and difficult challenge. It is in particular relevant for sentiment analysis, where singly labeled crowdsourced datasets are prevalent. This is due to data from the social web which is annotated by the data creators themselves, e.g., rating reviewers or categorizing image uploaders. This might further include multi-media contents such as audio, video, images, and other forms of texts. While the outlook for such forms of data is promising, end-to-end approaches have not yet been fully explored on these types of crowdsourcing applications.

%The paper at hand proposes an improvement to existing end-to-end approaches tailored for crowdsourcing.
%this scenario. End-to-end methods are a strong research trend and many of such models are the state of the art in several domains (\cite{oord2016wavenet,ma2016endtoend,borocs2018nlp}). This type of methods generally requires little data preprocessing if any at all and thereby imply small amounts of manual effort other than building the model itself. 
%As we show, end-to-end approaches have potential for sentiment analysis tasks and other domains where data is annotated by the data creators themselves, e.g., rating reviewers or categorizing image uploaders.
%, is taken from the web in large amounts. 
%This might further include multi-media contents such as audio, video, images, and other forms of texts. While the outlook for such forms of data is promising, end-to-end approaches have not yet been fully explored for these types of crowdsourcing applications.

With these benefits in mind, we propose a neural network model tailored for such data with singly labeled crowdsourced annotations. It computes a latent truth for each sample and the correct bias of every annotator while also considering input feature distribution during training. We modify the loss function such that \textit{the annotator bias converges towards the actual confusion matrix of the regarding annotator and thus models the annotator biases correctly}. 
This is novel, as previous methods either require a multi-labeled crowdsourcing setting \cite{Dawid:Skene:79,conf/naacl/HovyBVH13} or do not produce a correct annotator bias during training which would equal the confusion matrix, see \citet[figure 5]{conf/eccv/ZengSC18} and \citet[figure 3]{conf/aaai/RodriguesP18}. 
A correct annotator- or annotator-group bias, however, is necessary to derive correct conclusions about the respective annotator behavior. This is especially important for highly unreliable annotators who label a high number of samples randomly -- a setting, in which our proposed approach maintains its correctness, too.

Our contributions are as follows. We describe the corresponding state-of-the-art for crowdsourcing algorithms and tasks in section \ref{sec:related-work}. Our neural network model method for end-to-end crowdsourcing modeling is explained in section \ref{sec:method}, which includes a mathematical explanation that our linear bias modeling approach yields the actual confusion matrices. The experiments in section \ref{sec:experiments} underline our proof, show that the model handles annotator bias correctly as opposed to previous models, and demonstrate how the approach impacts classification.

%- list advantages of e2e
%  - e2e is a strong research trend
%  - e2e models are the state of the art in many domains (WaveNet, NLP-Cube \cite{borocs2018nlp}, End-to-end Sequence Labeling via Bi-directional LSTM-CNNs-CRF \cite{ma2016endtoend})
%  - little pre-processing necessary if any at all, so small manual effort apart from building the model itself
%  - consequently, a lot of potential
%- not fully explored for crowdsourcing applications
%- for singly labeled crowdsourcing, how to find the biases of annotators or annotator groups? This is a typical problem for sentiment analysis (→ reviewer annotations) which is a NLP domain. It might appear helpful for some domains to get rid of undesired biases regarding race, gender, age, and other, as has already been shown \cite{tripadvisor}.

%https://arxiv.org/abs/1805.04508

%- single label issue has also potential impact for other domains where data is taken from the web in large amounts which is annotated only by the creators. This might include multi-media contents such as audio, video, images, and other forms of texts which are categorized. \cite{standfordAIblog}

%- this is why we propose an e2e approach to crowdsourcing solving the denoted issues. 

%- further, tackling noisy labels coming from various annotators, annotator groups, or weak classifiers end-to-end also informs potential applications for weak supervision

\section{Related Work} \label{sec:related-work}

% TODO: RELATED WORK: BIAS CORRECTION IN ML!!!

\subsection{Crowdsourcing Algorithms}\label{sec:rel}

\textit{Problem definition.~} The need for data in the growing research areas of machine learning has given rise to the generalized use of crowdsourcing. 
This method of data collection increases the amount of data, saves time and money but comes at the potential cost of data quality. 
One of the key metrics of data quality is annotator reliability, which can be affected by various factors. For instance, the lack of rater accountability can entail spamming. \textit{Spammers} are annotators that assign labels randomly and significantly reduce the quality of the data. 
\citet{journals/jmlr/RaykarY12} and \citet{conf/naacl/HovyBVH13} addressed this issue by detecting spammers based on rater trustworthiness and the SpEM algorithm. 
However, spammers are not the only source of label inconsistencies. The varied personal backgrounds of crowd workers often lead to \textit{annotator biases} that affect the overall accuracy of the models. Several works have previously ranked crowd workers \cite{conf/naacl/HovyBVH13,conf/nips/WhitehillRWBM09,journals/jmlr/YanRFSVBMD10}, clustered annotators \cite{conf/acllaw/PeldszusS13}, captured sources of bias \cite{conf/nips/WauthierJ11} or modeled the varying difficulty of the annotation tasks \cite{Carpenter08multilevelbayesian,conf/nips/WhitehillRWBM09,10.5555/2997046.2997166} allowing for the elimination of unreliable labels and the improvement of the model predictions. 
%Other characteristics, benefits, and drawbacks of crowdsourcing have been explored by various authors (\cite{1613751,4562953}).

%\noindent
\textit{Ground truth estimation.~} One common challenge in crowdsourced datasets is the ground truth estimation. When an instance has been annotated multiple times, a simple yet effective technique is to implement majority voting or an extension thereof \cite{conf/nips/TianZ15,journals/jmlr/YanRFSVBMD10}. More sophisticated methods focus on modeling label uncertainty \cite{spiegelhalter1983analysis} or implementing bias correction \cite{1613751,journals/corr/abs-1906-01251}. These techniques are commonly used for NLP applications or computer vision tasks \cite{conf/nips/SmythFBPB94,journals/corr/abs-1906-01251}. Most of these methods for inferring the ground truth labels use variations of the EM algorithm by \citet{Dawid:Skene:79}, which estimates annotator biases and latent labels in turns. We use its recent extension called the \textit{Fast Dawid-Skene} algorithm \cite{sinha2018fast}. 
%FDS can be viewed as a 'hard' version of EM where the indicator variables per instance are binary (one-hot-encoded). This leads to faster convergence and good scalability in large datasets.

%\noindent
\textit{End-to-end approaches.~} The Dawid-Skene algorithm models the raters' \textit{abilities} as respective bias matrices. Similar examples include GLAD \cite{conf/nips/WhitehillRWBM09} or MACE \cite{conf/naacl/HovyBVH13}, which infer true labels as well as labeler expertise and sample difficulty. These approaches infer the ground truth only from the labels and do not consider the input features. \textit{End-to-end approaches} learn a latent truth, annotator information, and feature distribution jointly during actual model training \cite{conf/eccv/ZengSC18,khetan2017learning,conf/aaai/RodriguesP18}. Some works use the EM algorithm \cite{conf/icml/RaykarYZJFVBM09}, e.g., to learn sample difficulties, annotator representations and ground truth estimates \cite{platanios2020learning}. However, the EM algorithm has drawbacks, namely that it can be unstable and more expensive to train \cite{chu2020learning}. 
LTNet models imperfect annotations derived from various image datasets using a single latent truth neural network and dataset-specific bias matrices \cite{conf/eccv/ZengSC18}.
A similar approach is used for crowdsourcing, representing annotator bias by confusion matrix estimates \cite{conf/aaai/RodriguesP18}.
%by training two models on two different datasets and then using the datasets with the opposite models to obtain what they call pseudo-annotations 
%This is the base model of their architecture, and it is different from ours. 
%Both approaches are lacking a proper analysis regarding if the annotator bias is modeled correctly as mismatch between latent truth and annotations, and 
Both approaches show a mismatch between the bias and how it is modeled, see \citet[figure 5]{conf/eccv/ZengSC18} and \citet[figure 3]{conf/aaai/RodriguesP18}.
We adapt the LTNet architecture (see section \ref{sec:method}), as it can be used to model crowd annotators on singly labeled sentiment analysis, which, to our knowledge, is not done yet in the context of annotator bias modeling. Recent works about noisy labeling in sentiment analysis do not consider annotator bias \cite{journals/corr/abs-1909-00124}.

%Another important trend in crowdsourcing is end-to-end learning. Instead of resolving multiple labels into one label to train a classifier with, a probabilistic model is trained directly on the observed labels. Contrary to the Dawid-Skene or MACE algorithm, the model also takes the input into account. Most of the time this means representing the ground truth inside the model and then mapping it to the observed labels. Therefore, end-to-end learning is a combination of inferring the ground truth and modeling the annotator bias for debiasing purposes. In this fashion, \cite{conf/eccv/ZengSC18} train a neural network to infer the ground truth. Annotators and their bias are represented by confusion matrices similar to the Dawid-Skene algorithm. Both network and matrices are optimized jointly on the observed labels with back-propagation. This thesis examines a version of this approach, which is formally introduced in section \ref{sec:IPA2LT}.\\ 

\subsection{Crowdsourced Sentiment Datasets}

\textit{Sentiment and Emotion.~} Many works use the terms \textit{sentiment} and \textit{emotion} interchangeably \cite{demszky2020goemotions,kossaifi2019sewa}, whereas sentiment is directed towards an entity \cite{6797872} but emotion not necessarily. Both can be mapped to valence, which is the affective quality of goodness (high) or badness (low). Since emotion recognition often lacks annotated data, crowdsourced sentiment annotations can be beneficial \cite{1613751}.

%While both refer to reactions in the body and mind caused by social, biological, and cognitive factors, they are not identical. \cite{6797872} refers to sentiment as a latent factor that is constant, can be shaped by knowledge or previous experience, and is directed towards an object or an entity. Examples of sentiment are patriotism or friendship. On the other hand, emotion refers to the physical, physiological or psychological experience. It is briefer than a sentiment and not always directed towards an object. Arousal or depression are examples of emotion. 

%\noindent
\textit{Multi-Labeled Crowdsourced Datasets.~}
Crowdsourced datasets, such as, Google GoEmotion \cite{demszky2020goemotions} and the SEWA database \cite{kossaifi2019sewa}, usually contain multiple labels per sample and require their aggregation using ground truth estimation. Multi-labeled datasets are preferable to singly labeled ones on limited data. \citet{1613751} proved that many non-expert annotators give a better performance than a few expert annotators and are cheaper in comparison.

%\noindent
\textit{Singly Labeled Crowdsourced Datasets.~}
Singly labeled datasets are an option given a fixed budget and unlimited data. \citet{khetan2017learning} showed that it is possible to model worker quality with single labels even when the annotations are made by non-experts. Thus, multiple annotations can not only be redundant but come at the expense of fewer labeled samples. For singly labeled data, it can be distinguished between reviewer annotators and external annotators. Reviewer annotators rate samples they created themselves. It is common in forums for product and opinion reviews where a review is accompanied by a rating. As an example of this, we utilized the TripAdvisor dataset \cite{thelwall2018gender}. Further candidates are the Amazon review dataset \cite{ni2019justifying}, the Large Movie Review Dataset \cite{maas-EtAl:2011:ACL-HLT2011}, and many more comprising sentiment. External annotators annotate samples they have not created. Experts are needed for complex annotation tasks requiring domain knowledge. 
%Examples are the aspect-based sentiment analysis corpora from SemEval 2014, 2015 and 2016 \cite{semeval2014task4,semeval2015task12,pontiki2016semeval}. 
These are not crowdsourced, since the number of annotators is small and fixed. 
More common are external non-experts. \citet{1613751} showed that multi-labeled datasets annotated by non-expert improve performance. \citet{khetan2017learning} showed that it also performs well in the singly labeled case. Thus, datasets made of singly labeled non-expert annotations can be cheaper, faster, and obtain performances comparable to those comprised of different types of annotations. Our organic dataset is annotated accordingly, see section \ref{sec:organic_data}.

%\cite{danner2020combining, danner2020using}. 
%This dataset for aspect-based sentiment analysis is crowdsourced and labeled by external non-expert annotators. Every sample is annotated by a single annotator; hence it is a singly labeled dataset. However, the annotation task is a multi-class problem where each annotator potentially assigns multiple labels to the same sample. This is not to be confused with a crowdsourced multi-labeled dataset where multiple annotators label each sample. To our knowledge, this is the first dataset to approach aspect-based sentiment classification using this form of annotation. The organic dataset is described in section \ref{sec:organic_data}.

\begin{figure*}[t]
\centerline{\includegraphics[width=\textwidth]{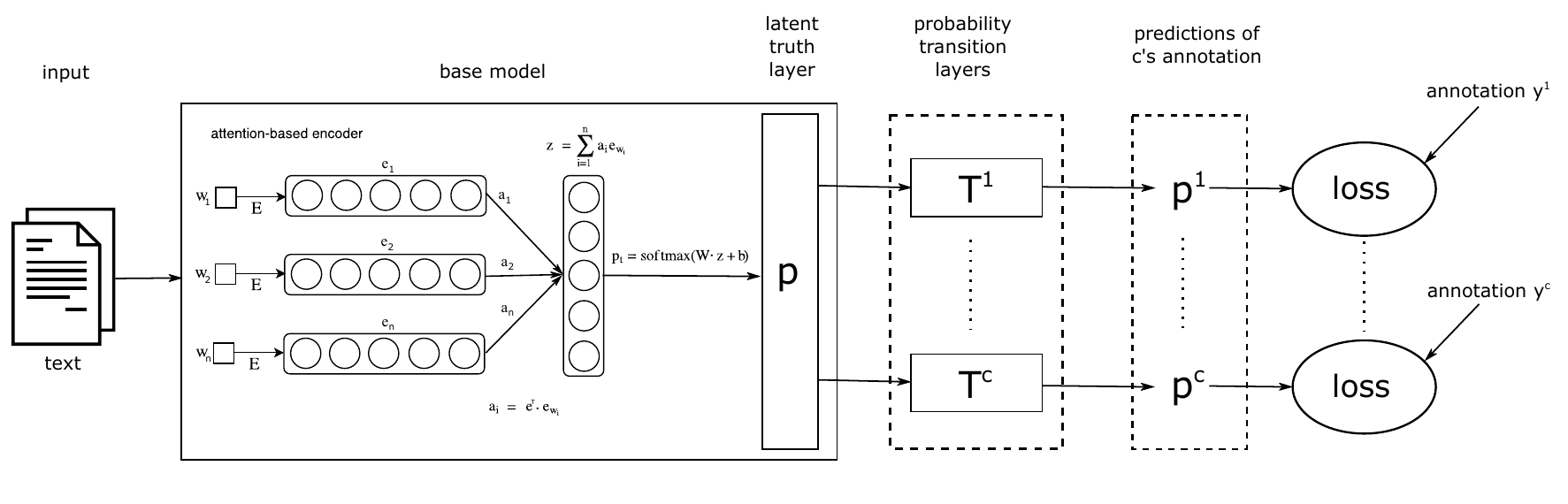}}
\caption{Architecture of the end-to-end trainable LTNet \cite{conf/eccv/ZengSC18}. The base model is a simple attention model with a single trainable attention vector $e$ and linear layer with parameters $W$ and $b$. The transition matrices $T^c$ are the bias matrices from the annotators $c$. \textquote[\citealp{conf/eccv/ZengSC18}]{Each row of the transition matrix T is constrained to be summed to 1}. The base model is inspired by ABAE \cite{he-etal-2017-unsupervised}.}
\label{fig:model}
\end{figure*}

% source in literature?

%\newpage
\section{Methodology} \label{sec:method}

\subsection{Basic Modeling Architecture}
The model choice is determined by the fact that some of our datasets are small. Thus, the model should have only few trainable parameters to avoid overfitting. We utilize a simple attention mechanism, as it is comon for NLP applications. The input words $w_j$ are mapped to their word embeddings $e_{w_j} \in \mathds{R}^{D}$ with $j=1,...,S$, and $S$ being the input sequence length and $D$ the dimensionality of the input word vectors. These are GloVe embeddings of 50 dimensions pre-trained on 6B English tokens of the "Wikipedia 2014 + Gigaword 5" dataset \cite{pennington2014glove}. Then, it computes the attention $a_i$ of each word using the trainable attention vector $e \in \mathds{R}^{D}$ via $a_j = e \cdot e_{w_j}$.
%In accordance with the continuous bag-of-words model \cite{mikolov2013efficient}, 
It takes the accordingly weighted average $z_n = \sum_{i=1}^{S} a_i \cdot e_{w_i}$ of the word vectors with $n$ denoting the $n$-th sample or input text. 

Finally, the classification head is the sigmoid of a simple linear layer $p_n = \text{softmax}(W \cdot z_n + b)$, with $W \in \mathds{R}^{L \times D}$ and $b \in \mathds{R}$ as the weights of the model. We refer to this last layer and to $p_n$ as \textit{latent truth layer} or \textit{latent truth}.

\subsection{End-to-End Crowdsourcing Model} 
\label{sec:ltnet}

On top of the basic modeling architecture, the biases of the annotators are modeled as seen in figure \ref{fig:model}. The theory is explained by \citet{conf/eccv/ZengSC18} as follows:

\begin{displayquote}
The labeling preference bias of different annotators cause inconsistent annotations. Each annotator has a coder-specific bias in assigning the samples to some categories.
Mathematically speaking, let $\mathcal{X} = \{ x_1, \dots , x_N \}$ denote the data, $y^c = [ y^c_1 , \dots , y^c_N ]$ the regarding annotations by coder $c$. Inconsistent annotations assume that $P(y^c_n | x_n) \neq P(y^{\hat{c}}_n | x_n) , \forall x_n \in \mathcal{X}, c \neq \hat{c}$, where $P(y^i_n | x_n)$ denotes the probability distribution that coder $c$ annotates sample $x_n$. 

LTNet assumes that each sample $x_n$ has a latent truth $y_n$. Without the loss of generality, let us suppose that LTNet classifies $x_n$ into the category $i$ with probability $P(y_n = i|x_n; \Theta)$, where $\Theta$ denotes the network parameters. If $x_n$ has a ground truth of $i$, coder $c$ has an opportunity of $\tau^c_{ij} = P(y^c_n = j | y_n = i)$ to annotate $x_n$ as $j$, where $y^c_n$ is the annotation of sample $x_n$ by coder $c$. Then, the sample $x_n$ is annotated as label $j$ by coder $c$ with a probability of 
$P(y^c_n = j | x_n; \Theta) = \sum^L_{i=1} P(y^c_n = j | y_n = i) P(y_n = i | x_n; \Theta)$, 
% \begin{flalign}
% & P(y^c_n = j | x_n; \Theta) = \nonumber \\
% & \sum^L_{i=1} P(y^c_n = j | y_n = i) P(y_n = i | x_n; \Theta), \nonumber
% \end{flalign}
where $L$ is the number of categories and \\
$ \sum^L_{j=1} P(y^c_n = j | y_n = i) = \sum^L_{j=1} \tau^c_{ij} = 1$. 

$T^c = [\tau^c_{ij}]_{L \times L}$ denotes the transition matrix (also referred to as annotator bias) with rows summed to $1$ while $\left[p_n\right]_i = P(y_n = i | x_n; \Theta)$ is modeled by the base network \cite{conf/eccv/ZengSC18}. We define $\left[p_n^c\right]_j = P(y^c_n = j | x_n; \Theta)$.
Given the annotations from $C$ different coders on the data, LTNet aims to maximize the log-likelihood of the observed annotations.
Therefore, parameters in LTNet are learned by minimizing the cross entropy loss of the predicted and observed annotations for each coder $c$.
\end{displayquote}

We represent the annotations and predictions as vectors of dimensionality $L$ such that $y^c_n$ is one-hot encoded and $p^c_n$ contains the probabilities for all class predictions of sample $n$. The cross entropy loss function is then defined as $- \sum_{n=1}^{C} \sum_{n=1}^{N} \log({p_n^c}^\intercal \cdot y_n^c )$.

% $$
% \min_{\Theta, \{T^1, \dots , T^C\}} 
% - \sum_{n=1}^{N} \log({p_n^c}^\intercal \cdot y_n^c ),
% $$

%Given the annotations from $C$ different coders on data $\mathcal{X}$, LTNet aims to maximize the loglikelihood of the observed annotations as
%\( \max_{\Theta, T^1, \dots , T^C}  
%\log \left( P(y^1, \newline y^2, \dots, y^C | \mathcal{X} ; \Theta) \right) \).

%It was already shown that finding the global optimum for the transition matrices $T^1, \dots T^c$ and the optimal parameter $\Theta$ is NP hard \cite{conf/eccv/ZengSC18}. Instead, the parameters in LTNet are approximated by minimizing the loss of the predicted and observed annotations for each coder.

\begin{figure*}[t]
\centering
\includegraphics[width=\textwidth]{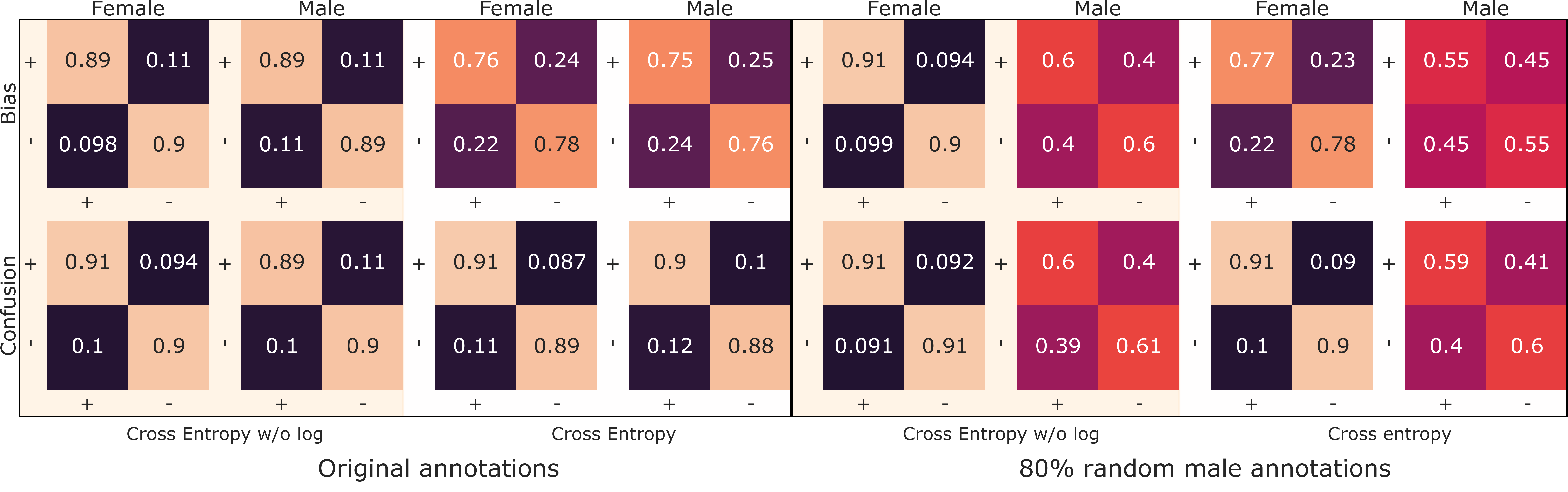}
\caption{
Male and female bias (top) and confusion (bottom) matrices which are trained using cross entropy loss with and without logarithm in two different settings. The left side has only the original annotations, whereas the right side has 80\% random male labels.
}
\label{fig:bias_convergence}
\end{figure*}

\subsection{The Effect of Logarithm Removal on Cross Entropy} 
\label{sec:proof}

%\subsection{NLLWO-Modification of the Loss Function}

%We propose that removing the logarithm from the cross entropy loss gives an advantage to find meaningful bias matrices. 
%We are not only interested in training an optimal model to make good predictions, but we especially pay attention to the correctness of the transition matrices as we want to use them to describe and effectively get rid of annotation bias. 

%We propose to omit the logarithm function from the cross entropy loss, to obtain what we define as the \textit{negative log-likelihood without logarithm} (NLLWO) loss. 

The logarithm in the cross entropy formula leads to an exponential increase in the loss for false negative predictions, i.e., when the predicted probability $\left[p^c_n\right]_i$ for a ground truth class $i$ is close to $0$ and $\left[y^c_n\right]_i$ is $1$. This increase can be helpful in conditions with numerical underflow, but at the same time this introduces a disproportionate high loss
%and thus \textit{class bias} 
of the other class due to constantly misclassified items. 
This happens in crowdsourcing, for example, when one annotator is a spammer assigning a high degree of random annotations, which in turn leads to a disproportionally higher loss caused by that annotator's many indistinguishable false negative annotations. Consequentially, the bias matrix of that annotator would be biased towards the false classes. Moreover, this annotator would cause overall more loss than other annotators, which can harm the model training for layers which are shared among all annotators, e.g., the latent truth layer when it is actually trained. 

By omitting the log function, these effects are removed and all annotators and datapoints contribute with the same weight to the overall gradient and to the trainable annotator bias matrices, independent of the annotator and his respective annotation behavior. 
As a consequence, the annotator matrices are capable of modeling the real annotator bias, which is the mismatch between an annotation $y^c_n$ of coder $c$ and the latent truth prediction $p_n$. If $p_n$ is one-hot encoded, this results to the according classification ratios of samples and is equal to the confusion matrix, without an algorithmically encoded bias towards a certain group of items. This is shown mathematically in the following, where it is assumed that the base network is fixed, i.e., backpropagation is performed through the bias matrices and stops at the latent truth layer.

We define $N = \sum_{k=1}^L N_k$ as the number of all samples and $N_k$ of class $k = 1, ..., L$. $L$ is the number of classes, $T^c = [\tau^c_{ij}]_{L \times L}$ the bias matrix of coder $c$, $p_n$ the latent truth vector of sample $n=1,...,N$, and $p_n^c$ the annotator prediction. $p_{km}$ is the latent truth of the $m$-th sample of class $k$ with $m=1,..., N_k$, same for $x_{km}$ and $y_{km}^c$. The loss without logarithm is

\begin{align*}
\boldsymbol{O} &= - \sum_{n=1}^{N} {p_n^c}^\intercal \cdot y_n^c \\
  &= - \sum_{k=1}^{L} \sum_{m=1}^{N_k} p_{km}^\intercal \cdot T^c \cdot y_{km}^c \\
  &= - \sum_{k=1}^{L} \sum_{m=1}^{N_k} p_{km}^\intercal \cdot 
     \begin{pmatrix}
     \tau^{c}_{1k} \\
     \vdots \\
     \tau^{c}_{Lk} 
     \end{pmatrix} \\
  &= \sum_{k=1}^{L} \sum_{m=1}^{N_k} \sum_{h=1}^{L} 
     - \left[p_{km}\right]_h \cdot \tau^c_{hk}
\end{align*}

%Now, let the learning rate be $\alpha$, the number of epochs $E$ and the starting values $\left(\tau^{c}_{1k}\right)_0$ and $\left(\tau^{c}_{2k}\right)_0$ respectively. The bias parameters $\tau^c_{1k}$ and $\tau^c_{2k}$ change according to

Apparently, the derivation step between the second and third line would not work if there would be the logarithm from the standard cross entropy.
Now, let the learning rate be $\alpha$, the number of epochs $E$ and the starting values of the initialized bias matrix $\left(\tau^{c}_{lh}\right)_0$. The bias parameters $\tau^c_{lh}$ of the bias matrix $T^c$ are updated according to

\begin{align*}
\left(\tau^{c}_{hk}\right)_E 
  &= \left( \tau^{c}_{hk} \right)_0 
     + \sum_{i=1}^E \alpha 
     \left( \frac{\partial \boldsymbol{O}}{\partial \tau^c_{hk}} \right)_i \\
  &= \left( \tau^{c}_{hk}\right)_0
     + \sum_{i=1}^E \alpha 
     \left[ \sum_{m=1}^{N_k} - \left[p_{km}\right]_h \right]_i \\
  &= \left( \tau^{c}_{hk} \right)_0 
     - \alpha E 
     \underbrace{\sum_{m=1}^{N_k} \left[p_{km}\right]_h}_{=: Z_{hk}}
%\left(\tau^{c}_{2k}\right)_E
%  &= \left(\tau^{c}_{2k}\right)_0 
%     + \sum_{i=1}^E \alpha 
%     \left(\frac{\partial C}{\partial \tau^c_{2k}}\right)_i \\
%  &= \left(\tau^{c}_{2k}\right)_0 
%     - \alpha E 
%     \underbrace{\sum_{m=1}^{N_k} {p_{km}}_2}_{=: Z_{2k}}.
\end{align*}

For sufficiently large $E$ the starting values $\left(\tau^{c}_{hk}\right)_0$  become infinitesimally small in comparison to the second additive term and thus negligible. As we are normalizing the rows of $(T^c)_E$ after training so that the bias fulfills our probability constraint defined in section \ref{sec:ltnet}, the linear factor $-\alpha E$ is canceled out, too. Thus, the bias matrix $T^c$ results in the row normalized version of $[Z_{hk}]_{L \times L}$. $Z_{hk}$ is the sum of the latent truth probabilities for class $h$ on all samples of a ground truth class $k$. If we assume that the latent truth is one hot encoded, $[Z_{hk}]_{L \times L}$ equals to the confusion matrix, of which the $k$-th column sums up to the number of samples in class $k$: $\sum_{h=1}^L Z_{hk} = \sum_{h=1}^L \sum_{m=1}^{N_k} \left[p_{km}\right]_h = \sum_{m=1}^{N_k} 1 = N_k$.

\section{Experiments} \label{sec:experiments}

\subsection{Bias Convergence}
\label{sec:bias-convergence-experiment}

%and with stochastic gradient descent 

The following experiment compares how training with and without the logarithm in the cross entropy loss affects the LTNet bias matrices empirically. The mathematical explanations in section \ref{sec:proof} suggest that the logarithm removal from cross entropy leads to an annotator bias matrix identical to the confusion matrix, which would not be the case for the normal cross entropy.

\textbf{Experiment Description.~}
For the data, we use the TripAdvisor dataset from Thelwall et al. consisting of $11,900$ English consumer reviews about hotels from male and female reviewers plus their self-assigned sentiment ratings \cite{thelwall2018gender}. We use the gender information to split the data into two annotator groups, male and female, from which we model each one with a corresponding bias matrix. We exclude neutral ratings and binarize the rest to be either positive or negative. As the dataset is by default completely balanced regarding gender and sentiment at each rating level, it is a natural candidate for correct bias approximation.
Throughout our experiments, we use 70\% of the obtained data as training, 20\% as validation and the 10\% remaining as test sets. 

%Reviewer genders are annotated from their names, which is claimed to be over 90\% accurate. 

%The proposed LTNet was implemented under the framework of PyTorch.
Similar to the explanation in \ref{sec:proof}, the base model with its latent truth predictions is pre-trained on all samples and then frozen when the bias matrices are trained. The stochastic gradient descent method is used to optimize the parameters, as other wide-spread optimizers, such as Adam and AdaGrad (the latter introduced that feature first), introduce an -- in our case undesired -- bias towards certain directions in the gradient space, namely by using the previous learning steps to increase or decrease the weights along dimensions with larger or smaller gradients \cite{journals/corr/KingmaB14}. 
%We conduct experiments to compare the cross entropy and the NLLWO loss. We use two different settings of the TripAdvisor dataset to demonstrate what happens if the latent truth layer is more and less certain. 
For all four sub-experiments, we train the base models with varying hyperparameters and pick the best based on accuracy. We train the transition matrices 50 times with different learning rates from the interval $[1\mathrm{e}{-6}, 1\mathrm{e}{-3}]$. The batch size is 64. 
In addition to a normal training setting, we add random annotations to 80\% of the instances annotated by male subjects, such that ~40\% from them are wrongly annotated. This results in four models: with and without logarithm in the cross entropy, with and without random male annotations, each time respectively with two annotator group matrices, male and female -- see figure \ref{fig:bias_convergence}.

\textbf{Results.~}
%The classification accuracy of both losses is almost identical with maxima of $0.8978$ and $0.8973$ accuracy with and without logarithm in the no noise experiment. 
The bias matrices of the models with the best accuracy are picked and presented in figure \ref{fig:bias_convergence} in the top row. The corresponding confusion matrices depict the mismatch between latent truth predictions and annotator-group labels in the bottom row. The bias matrices trained without logarithm in the cross entropy are almost identical to the confusion matrices in all cases, which never holds for the normal cross entropy. This confirms our mathematically justified hypothesis given in section \ref{sec:proof} that the logarithm removal from cross entropy leads to a correctly end-to-end-trained bias. In this context, it is relevant that the related work shows the same mismatch between bias and confusion matrix when applying cross entropy loss without explaining nor tackling this difference, see \citet[figure 5]{conf/eccv/ZengSC18} and \citet[figure 3]{conf/aaai/RodriguesP18}. 

It is worth mentioning for the 80\% random male annotations that these are correctly modeled without cross entropy, too, as opposed to normal cross entropy. If the goal is to model the annotator bias correctly in an end-to-end manner, this might be considered as particularly useful to analyze annotator behavior, e.g., spammer detection, later on.

Finally, we report how much variation the bias matrices show during training for cross entropy with and without logarithm. As mentioned in the experiment description, we trained each model $50$ times. The elements of the resulting bias matrices with standard cross entropy have on average $7.7\%$ standard deviation compared to $2.8\%$ without logarithm. It can be concluded that the bias produced by standard cross entropy is less stable during training, which raises questions about the overall reliability of its outcome.

In summary, the observations confirm our assumptions that cross entropy without logarithm captures annotator bias correctly in contrast to standard cross entropy. This carries the potential to detect spammer annotators and leads to an overall more stable training.

%A possible reason for this is the non-linearity of cross entropy that puts much more weight on misclassified items than others in the gradient space. It gives more weight on misclassified items leading to bias matrices which are more difficult to interpret. 

\subsection{Ground Truth Estimation}

In the following paragraphs, we demonstrate how to estimate the ground truth based on the latent truth from LTNet. 
This is then compared to two other kinds of ground truth estimates.
All of them can be applied in a single label crowdsourcing setting. 

The Dawid-Skene algorithm \cite{sinha2018fast} is a common approach to calculate a ground truth in crowdsourcing settings where there are multiple annotations given on each sample. This method is, for instance, comparable to majority voting, which tends to give similar results for ground truth estimation. However, in single label crowdsourcing settings, these approaches are not feasible. Under single label conditions, the Dawid-Skene ground truth estimates equal to the single label annotations. 

This is given by \citet[formula 1]{sinha2018fast} in the expectation step, where the probability for a class $k \in {1,2,...,L }$ given the annotations is defined as

\begin{align*}
P ( Y_n = k | k_{n_1} , k_{n_2} , ..., k_{n_L} ) = \\
\frac{
    \left( \prod_{c=1}^C P (k_{n_c} | Y_n = k) \right) \cdot P (Y_n = k)
}{
    \sum^L_{k=1} \left( \prod_{c=1}^C P (k_{n_c} | Y_n = k) \right) \cdot P (Y_n = k)
}.
\end{align*}

Here, $n$ is the sample to be estimated, $C$ the number of annotators for that sample, 
${n_1, n_2, ..., n_C}$ the set of annotators who labeled this sample, 
${k_{n_1} ,k_{n_2} ,...,k_{n_C}}$ the set of annotation choices chosen by these $C$ participants for sample n, 
and $Y_n$ the correct (or aggregated) label to be estimated for the sample $n$ \cite{sinha2018fast}. 

In the single label case $C$ equals to 1, which reduces the formula to 
$P ( Y_n = k | k_{n_1} , k_{n_2} , ..., k_{n_C} ) = P ( Y_n = k | k_{n_1})$. This in turn equals to $1$ if $k$ is the assigned class label to sample $n$ by annotator $n_1$, or $0$ otherwise. In other words, if there is only one annotation per sample, this annotation defines the ground truth. Since different annotators do not assign labels on the same samples, there is also no way to model mutual dependencies of each other.

LTNet, however, provides estimates for all variables from this formula. $P (Y_n = k)$ is the prior and is approximated by the latent truth probability for class $k$ of sample $n$. $P (k_{n_c} | Y_n = k)$ is the probability that, assuming $k$ would be the given class, sample $n$ is labeled as $k_{n_c}$ by annotator $n_c$. This equals to $\tau^c_{k_{n_c},k}$, i.e., the entries of the LTNet bias matrix $T^c$ of annotator $c$.

Eventually, the LTNet ground truth can be derived by choosing $k$ such that the probability $P ( Y_n = k | k_{n_1} , ... )$ is maximized:

\begin{align*}
k_{\mathrm{ground\, truth}} = \underset{k}{\mathrm{arg\,max}} \, P ( Y_n = k | k_{n_1} , ... ).
\end{align*}

We will leverage this formula to derive and evaluate the ground truth generated by LTNet.

\paragraph{Experiment}
We calculate the LTNet ground truth according to the previous formula on the organic dataset, a singly labeled crowdsourcing dataset, which is described in Section \ref{sec:organic_data}. To demonstrate the feasibility and the soundness of the approach, we compare it with two other ways of deriving a ground truth. Firstly, we apply the fast Dawid-Skene algorithm on the annotator-wise class predictions from the LTNet model. Secondly, we train a base network on all annotations while ignoring which annotator annotated which samples. Eventually, we compare the ground truth estimates of all three methods by calculating Cohen's kappa coefficient \cite{cohen1960coefficient}, which is a commonly used standard to analyze correspondence of annotations between two annotators or pseudo annotators. The training procedures and the dataset are identical to the ones from the classification experiments in Section \ref{sec:classification}. 

\paragraph{Results}

As can be seen on Table \ref{tab:cohen-kappas}, the three ground truth estimators are all highly correlated to each other, since the minimal Cohen's kappa score is $0.98$. Apparently, there are only minor differences in the ground truth estimates, if any at all. 
Thus, it appears that the ground truths generated by the utilized methods are mostly identical. Especially, the LTNet and Dawid-Skene ground truths are highly correlated with a kappa of \( 99\% \). The base model, which is completely unaware of which annotator labeled which sample, is slightly more distant with kappas between \( 98\% \) -- \( 99\% \). So with respect to the ground truth itself, we do not see a specific benefit of any method, since they are almost identical.

However, it must be noted that LTNet additionally produces correct bias matrices of every annotator during model training, which is not the case for the base model. Correct biases have the potential to help improving model performance by analyzing which annotators tend to be more problematic and weighting them accordingly. 
%Even though it is not investigated here, it shall not be left unmentioned that this could also be leveraged by the Dawid Skene method based on LTNet predictions.

\begin{table}[t]
\centering
\begin{tabular}{l|ccc} 
& Dawid &  & Basic \\ 
Ground truths & Skene & LTNet & Model \\ \hline 
Dawid Skene & $1.0000$ & $0.9905$ & $0.9832$ \\ 
LTNet & $0.9905$ & $1.0000$ & $0.9918$ \\ 
Base Model  & $0.9832$ & $0.9918$ & $1.0000$ \\ 
\end{tabular}
\caption{Cohen's kappa scores between three different ground truth estimation methods applied on the singly labeled crowdsourced organic dataset.}
\label{tab:cohen-kappas}
\end{table}

\subsection{Classification} \label{sec:classification}

We conduct classification comparing LTNet in different configurations on three datasets with crowdsourced sentiment annotations to discuss the potential related benefits and drawbacks of our proposed loss modification.

\textbf{Emotion Dataset.~} \label{sec:data}
The emotion dataset consists of 100 headlines and their ratings for valence by multiple paid Amazon Mechanical Turk annotators \cite{1613751}. Each headline is annotated by 10 annotators, and each annotated several but not all headlines. We split the interval-based valence annotations to positive, neutral, or negative. Throughout our experiments, we used 70\% of the obtained data as training, 20\% as validation and 10\% as test sets. 

%\textbf{TripAdvisor Dataset.~}
%See section \ref{sec:bias-convergence-experiment}.
%
%\noindent
%\textbf{Emotion Dataset.~} 

%(crowdsourced bias)

%We could replace the Jurafsky emotion dataset by the Wikipedia toxicity dataset, as both are multi-labeled such that we can compare against the Dawid-Skene baseline. The emotion dataset still might be more suitable, as it has smaller inter-rater reliability. High reliability is already covered by the TripAdvisor dataset. There we already see that our algorithm does not bring accuracy improvements when we have high reliability. So, it would be a bit boring. Even if we are not improving on the emotion dataset or even get worse, we can point out the limitations of the approach. Anyway, Wikipedia toxicity is still a valid option.

%\noindent
\textbf{Organic Food Dataset.~} \label{sec:organic_data}
With this paper, we publish our dataset containing social media texts discussing organic food related topics.

\noindent
\textit{Source.~} The dataset was crawled in late 2017 from Quora, a social question-and-answer website. To retrieve relevant articles from the platform, the search terms "organic", "organic food", "organic agriculture", and "organic farming" are used. The texts are deemed relevant by a domain expert if articles and comments deal with organic food or agriculture and discuss the characteristics, advantages, and disadvantages of organic food production and consumption. 
%Typical articles that were found but are not relevant are recipes, product presentations, and stock market information. The remaining irrelevant posts were filtered out by a domain expert.
From the filtered data, 1,373 comments are chosen and 10,439 sentences annotated.

%by applying a binary naive Bayes classifier based on bag-of-words. A market research domain expert labeled 1000 randomly selected posts as either domain relevant or irrelevant. This was used as training data. Each document in the training and test set is composed by concatenating the question text, and the text of the first 100 comments. The output of the classification was shown to the domain expert and revised until satisfactory results were achieved. Using 10-fold cross validation the accuracy of the classifier eventually was 84.70\%.

\noindent
\textit{Annotation Scheme.~} Each sentence has sentiment (positive, negative, neutral) and entity, the sentiment target, annotated. We isolate sentiments expressed about organic against non-organic entities, whereas for classification only singly labeled samples annotated as organic entity are considered. Consumers discuss organic or non-organic products, farming practices, and companies.

%Attributes are not considered for sentiment analysis. 

%We additionally include opinions about genetic engineering. Regarding attributes, we find arguments such as price, quality, healthiness, trustworthiness, and environmental issues to be typical reasons why organic and non-organic is being referred to as something positive or negative. 

%With respect to attributes, we find relevant arguments such as price, quality, healthiness, trustworthiness, and environmental issues to be typical reasons why organic and non-organic is being referred to as something positive or negative.

%Let us consider an example sentence \textit{"I just want to say that organic husbandry is not all good or bad"}. As the topic of the sentence is about the organic food domain, it is self-explanatory that it would be categorized as \textit{domain relevant} when thinking of the first category. Considering the sentiment of the phrase, one can neither can say that it is positive nor that it is negative, and therefore it would be labeled with \textit{ambiguous/indifferent/no sentiment}. The entity of the sentence would be ranked with the option \textit{organic farmers} and the attribute as \textit{animal welfare}. 

%The entity of the sentence would be ranked with the option \textit{organic farmers} and the attribute as \textit{animal welfare}. 

\noindent
\textit{Annotation Procedure.~} The data is annotated by each of the 10 coders separately; it is divided into 10 batches of $1,000$ sentences for each annotator and none of these batches shared any sentences between each other. 
%No voting algorithm is used to decide the final gold label for a sentence. 
%Table \ref{tab:annotation-rule-ent} shows the eventual distribution of labels over all given annotations.
4616 sentences contain organic entities with 39\% neutral, 32\% positive, and 29\% negative sentiments.
After annotation, the data splits are 80\% training, 10\% validation, and 10\% test set. The data distribution over sentiments, entities, and attributes remains similar on all splits.

%\subsection{Training Modes}

%LTNet is trained using backpropagation but in two possible ways. Either the base model is pre-trained and the bias matrices on top are trained in a separate step afterwards, or both are trained jointly in the same step. We will refer to the former approach as \textit{fixed base}, as we keep the base model frozen during the training of the matrices, and to the latter as \textit{end-to-end} approach. In case of \textit{end-to-end}, we train a base model, add the transition matrices, and train and fine-tune everything together. The base model can be trained with any suitable loss function. For fair comparison in evaluation, we use the same base models to start within the case of the \textit{end-to-end} approach.

\textbf{Experiment Description.~}
%In this experiment, we investigate whether the LTNet architecture improves classification compared to the base network, as the former additionally captures the underlying bias using bias matrices whereas the latter does not.
The experiment is conducted on the TripAdvisor, organic, and emotion datasets introduced in section \ref{sec:data}. We compare the classification of the base network with three different LTNet configurations. Two of them are trained using cross entropy with and without logarithm. For the emotion dataset, we compute the bias matrices and the ground truth for the base model using the fast Dawid-Skene algorithm \cite{sinha2018fast}. This is possible for the emotion dataset, since each sample is annotated by several annotators.

%We used the end-to-end approach to train the models. 
We apply pre-training for each dataset by training several base models with different hyperparameters and pick the best based on accuracy. Eventually, we train the LTNet model on the crowdsourcing annotation targets by fine-tuning the best base model together with the bias matrices for the respective annotators. The bias matrices are initialized as row normalized identity matrices plus uniform noise around $0.1$. The models are trained 50 times with varying learning rates sampled from between $[1\mathrm{e}{-6}, 1\mathrm{e}{-3}]$. A batch size of 64 is used.

\begin{table}[t]
\small
\centering
%\begin{wraptable}{l}{67mm}
\setlength\tabcolsep{4.9pt}
\begin{tabular}{llrr}
\toprule
\textbf{Dataset} & \textbf{Model} & \textbf{F1 \%} & \textbf{Acc \%} \\
\midrule
\multirow{3}*{TripAdvisor} & Base Model & 88.92 &  88.91 \\
 & LTNet w/o log & \textbf{89.71} & \textbf{89.71} \\
 & LTNet & 89.39 & 89.39 \\
\hline
\multirow{3}*{Organic} & Base Model & 32.08 & 45.75 \\
 & LTNet w/o log & \textbf{44.71} & \textbf{50.54} \\
 & LTNet & 40.51 & 47.77 \\
\hline
\multirow{4}*{Emotion} & Base Model & 51.74 & 56.00 \\
 & LTNet w/o log & 58.15 & 63.00 \\
 & LTNet & \textbf{61.23} & \textbf{66.00} \\
 %& LTNet + Dawid-Skene & 52.45 &  \\
 & Base Model DS & 44.17 & 54.00 \\
\bottomrule
\end{tabular}
\caption{
Macro F1 scores and accuracy measured in the classification experiment. %NLLWO and CE mean negative log-likelihood without logarithm and cross entropy. DS means the base model is trained on the ground truth of the Dawid-Skene algorithm. The emotion dataset is small resulting in seemingly rounded accuracies.
}
\label{tab:classification_table}
\end{table}
%\end{wraptable}

\textbf{Results.~}
The classification results of the models are presented in table \ref{tab:classification_table} with their macro F1 score and accuracy as derived via predictions on the test sets. LTNet generally shows a significant classification advantage over the base model. On all three databases, LTNet approaches performed better on the test datasets. The LTNet improvement has a big delta of $11\% +/- 1\%$ when there is a low annotation reliability (organic and emotion datasets) and a small delta $<1\%$ with high reliability (TripAdvisor) \footnote{Unreliable means that the provided annotations have a low Cohen's kappa inter-rater reliability on the organic $51.09\%$ and emotion ($27.47\%$) dataset. On the organic dataset we prepared a separate data partition of 300 sentences annotated by all annotators for that purpose. For the TripAdvisor dataset, it is apparent that the correspondence of annotations between the two annotator groups (male and female) is high as can be seen in figure \ref{fig:bias_convergence} for cross entropy without logarithm.}. Apparently, model each annotator separately gives significant advantages.

%Even though there is no option to calculate the inter-rater reliability, as the male and female reviewers only annotated once their review, the male and female NLLWO bias matrices in figure \ref{fig:bias_convergence} look almost identical. This gender-specific annotator group bias is derived by applying NLLWO loss which means that it can be assumed to converge to the actual bias inherent to the data itself as shown in section \ref{sec:proof}. Thus, the NLLWO bias matrices imply that the male and female bias is almost equal, which means that the sentiments are annotated in a very similar manner, leading to high reliability. We conclude that the smaller the inter-rater reliability is, the more improvement is achieved by applying LTNet for the denoted crowdsourced sentiment analysis tasks. This effect is, theoretically, independent of the loss function.

Regarding the comparison between cross entropy (CE) loss with and without logarithm on LTNet, the removed logarithm shows better classification results on organic ($+3\%$) and TripAdvisor data ($+0.3\%$) and worse on the emotion dataset ($-3\%$). This means that on both of the singly labeled crowdsourcing datasets, the removal of the logarithm from the loss function leads to better predictions than the standard CE loss. On the multi-labeled emotion dataset, however, this does not appear to be beneficial. As this data has only a very small test set of 100 samples, it is not clear if this result is an artifact or not. Concluding, the log removal appears to be beneficial on large datasets, where the bias is correctly represented in the training and test data splits, such that it can be modeled correctly by the denoted approach. It shall be noted, that it is not clear if that observation would hold generally. We advice to run the same experiments multiple times on many more datasets to substantiate this finding.

%The latter, however, improves results for the multi-annotator labeled emotion dataset. 
%This supports the hypothesis that the removal of the logarithm from the loss function might be more suitable for singly labeled and CE loss for multi-labeled crowdsourcing annotations.

%The classification results on the emotion dataset might not be reliable as the whole dataset only consists out of $100$ datapoints overall. 

%In contrast, the organic dataset has overall $4592$ sentences annotated with respective sentiments, which gives a more representative result, not to speak about TripAdvisor with $19,000$ samples. Overall, the results show at least that NLLWO loss does not necessarily decrease the results and might improve them, e.g., when many unreliable annotations from different annotators with varying annotation behaviors are present. 

% especially advantageous with many annotators and highly incoherent annotations
%\balance

%Finally, the Dawid-Skene approach produces the worst results on the emotion dataset. Our explanation is that it is a step-by-step method, which means it separately captures the relations between the input and the latent truth, and the relations between the latent truth and the inconsistent annotations. This inflexibility and the small amount of data explain its poor results on the emotion task.

\section{Conclusion}

We showed the efficacy of LTNet for modeling crowdsourced data and the inherent bias accurately and robustly. The bias matrices produced by our modified LTNet improve such that they are more similar to the actual bias between the latent truth and ground truth. Moreover, the produced bias shows high robustness under very noisy conditions making the approach potentially usable outside of lab conditions. The latent truth, which is a hidden layer below all annotator biases, can be used for ground truth estimation in our single label crowdsourcing scenario, providing almost identical ground truth estimates as pseudo labeling. 
Classification on three crowdsourced datasets show that LTNet approaches outperfom naive approaches not considering each annotator separately.
The proposed log removal from the loss function showed better results on singly labeled crowdsourced datasets, but this observation needs further experiments to be substantiated.
%The proposed approach was not yet utilized for ground truth estimation, which might have the potential for improvements in that regard, as the input features are also taken into account here. 
%We think it is necessary to conduct additional experiments on more datasets to solidify our conclusions regarding classification, as the multi-labeled crowdsourcing use case is performed on a too small dataset. 
Furthermore, there might be many use cases to explore the approach on other tasks than sentiment analysis.
%There are presumably use cases which are potentially relevant for building unbiased and thus fairer classification systems, potentially with positive societal implications.

\balance

\bibliography{main}
\bibliographystyle{acl_natbib}

\appendix

\end{document}